%% file: colm2025_conference.tex
\pdfoutput=1

\documentclass{article} 
\usepackage[final]{colm2025_conference}

\usepackage{times}
\usepackage{latexsym}

\usepackage[T1]{fontenc}
\usepackage[utf8]{inputenc}
\usepackage{microtype}
\usepackage{inconsolata}
\usepackage{graphicx}
\usepackage{hyperref}
\usepackage{booktabs}
\usepackage{xspace}
\usepackage{makecell}
\usepackage{float}
\usepackage{multirow}
\usepackage{multicol}
\usepackage{natbib}
\usepackage{amsmath}
\usepackage{comment}
\usepackage{textcomp}
\usepackage{pifont}
\usepackage{acronym}

\usepackage{longtable}

\newcommand*{\flores}{\textsc{Flores}\xspace}
\newcommand*{\tweetlid}{\textsc{TweetLID}\xspace}
\newcommand*{\commonvoice}{\textsc{Common Voice}\xspace}
\newcommand*{\smol}{\textsc{SMOL}\xspace}
\newcommand*{\bloom}{\textsc{Bloom}\xspace}
\newcommand*{\csbench}{\textsc{Divers-Bench}\xspace}
\newcommand*{\diverscs}{\textsc{Divers-CS}\xspace}
\newcommand*{\diversmono}{\textsc{Divers-Mono}\xspace}

\input{full_language_eval_table_reordered}

\title{DIVERS-Bench: Evaluating Language Identification Across Domain Shifts and Code-Switching}

\author{%
Jessica Ojo$^1$\thanks{Equal contribution} \quad
Zina Kamel$^1$\footnotemark[1] \quad David Ifeoluwa Adelani$^{1,2}$\\
\footnotesize
$^1$Mila - Quebec AI Institute \& McGill University, $^2$ Canada CIFAR AI Chair \\
\footnotesize
\texttt{Correspondence:\{jessica.ojo, zina.kamel\}@mail.mcgill.ca}}

\begin{document}
\maketitle

\begin{abstract}
Language Identification (LID) is a core task in multilingual NLP, yet current systems often overfit to clean, monolingual data. This work introduces \csbench a comprehensive evaluation of state-of-the-art LID models across diverse domains - including speech transcripts, web text, social media texts, children’s stories, and code-switched text. Our findings reveal that while models achieve high accuracy on curated datasets, performance degrades sharply on noisy and informal inputs. We also introduce \diverscs, a diverse code-switching benchmark dataset spanning 10 language pairs, we show that existing models struggle to detect multiple languages within the same sentence. These results highlights the need for more robust and inclusive LID systems in real-world settings. \footnote{https://github.com/McGill-NLP/DIVERS-Bench}
\end{abstract}

\section{Introduction}
\vspace{-2mm}
Language identification (LID) which is the task of determining the language of a given text, is foundational in multilingual NLP, enabling applications such as translation, speech recognition, and content moderation \citep{burchell-etal-2023-open}. While modern LID systems report high accuracy on benchmark datasets and cover hundreds of languages \citep{goswami-etal-2024-native, Jauhiainen2019, kargaran-etal-2023-glotlid}, their real-world performance remains limited.

State-of-the-art LID models often evaluate on clean, high-resource, single-label data and struggle with noisy, informal, or code-switched inputs. Real-world text frequently includes slang, misspellings, and dialectal variation \citep{lid_wild}, yet evaluations remain dominated by curated datasets such as WiLI \citep{thoma2018wilibenchmarkdatasetwritten}, UDHR \citep{kargaran-etal-2023-glotlid}, and FLORES \citep{goyal-etal-2022-flores}, which fail to capture this complexity \cite{tokenlevel_lid}.

This study presents a comprehensive evaluation of LID robustness across three axes: domain diversity, language resource imbalance, and code-switched inputs. We focus on two questions: (1) How robust are current LID models to noisy and informal text? (2) How well do they handle code-switching across diverse language pairs?

We evaluate eight LID models on datasets spanning speech transcripts, social media, children stories, and translation corpora. We also introduce a diverse code-switching (CS) benchmark covering 10 language pairs across high- and low-resource settings, the most comprehensive code-switching dataset to date. Our evaluation framework provides a more inclusive view of LID performance across real-world conditions.

\section{Related Works}
\vspace{-2mm}

Language identification (LID) is a critical preprocessing step in multilingual NLP, supporting tasks such as translation, transcription, and parsing~\citep{Panich2015ComparisonOL}. Early LID systems relied on character n-gram features with classifiers like Naive Bayes and SVMs~\citep{baldwin-lui-2010-language}. With the rise of deep learning, scalable neural models have emerged. FastText-LID~\citep{joulin-etal-2017-bag}, a shallow n-gram-based model trained on Wikipedia, remains widely used. Its architecture has been extended in systems like GlotLID~\citep{kargaran-etal-2023-glotlid}, covering over 1,600 languages. Other notable models include CLD3 (LSTM-based), AfroLID~\citep{adebara-etal-2022-afrolid} for African languages, MaskLID~\citep{kargaran-etal-2024-masklid} for code-switching, and OpenLID~\citep{burchell-etal-2023-open}, which focuses on curated multilingual sources. Most LID models report strong performance on benchmarks such as WiLI-2018~\citep{thoma2018wilibenchmarkdatasetwritten}, Tatoeba~\citep{tiedemann-2020-tatoeba}, LangID-33~\citep{lui-baldwin-2012-langid}, and FLORES \cite{goyal-etal-2022-flores}. However, these datasets are typically clean, formal, and monolingual, and thus fail to reflect real-world variation, including noisy, informal, short, or mixed-language inputs~\citep{lid_wild, lid_web, dogruoz-etal-2021-survey}. Studies such as~\cite{tweetlid, 10.1007/s10579-012-9195-y} show that even state-of-the-art models degrade substantially on social media and microblogs, which often feature slang, emojis, dialects, and non-standard grammar.

In addition, real-world language use frequently involves the alternation between two or more languages within a single sentence or utterance referred to as CS which poses a significant challenge for existing LID systems~\citep{burchell-etal-2024-code, winata-etal-2023-decades}. While corpora like TongueSwitcher~\citep{sterner2023tongueswitcher} and CanVEC~\citep{nguyen-bryant-2020-canvec} address CS, they typically focus on narrow language pairs or domains, limiting generalizability. OpenLID has extended evaluation to a handful of CS scenarios, but broader coverage remains scarce. 

Our research addresses this gap by building upon and complementing existing work. We perform an extensive evaluation of SoTA LID models on a diverse collection of formal and informal datasets, including speech transcripts, web-crawled text, social media, children's stories, and professional translations. We also build on these efforts by introducing \diverscs a diverse CS benchmark covering 13 language pairs - spanning both high- and low-resource languages to provide a more inclusive evaluation of LID robustness in mixed-language settings.

\section{Evaluation Strategy}
\vspace{-3mm}
\subsection{LID Models}
\vspace{-2mm}
We evaluate eight LID models spanning a broad range of techniques, language coverage, and training regimes, as summarized in Table~\ref{tab:lid-models}. These include traditional models such as Google's Compact Language Detector 3 (CLD3), LangDetect, and LangID.py~\citep{lui-baldwin-2012-langid}, which rely on statistical, frequency-based, or rule-based methods. Data-driven approaches include FastText LID~\citep{joulin2016fasttext} and Franc, while recent large-scale systems such as GlotLID~\citep{kargaran-etal-2023-glotlid}, OpenLID~\citep{burchell-etal-2024-code}, and ConLID~\citep{foroutan2025conlidsupervisedcontrastivelearning} leverage multilingual corpora and deep learning. ConLID, in particular, employs supervised contrastive learning to improve robustness on out-of-domain and low-resource languages. Appendix~\ref{sec:model_descriptions} describes models evaluated.
\vspace{-2mm}

\subsection{\csbench: Datasets and Domains}
\vspace{-2mm}

We introduce \csbench a new benchmark covering several domains and real-world CS corpora.

To evaluate the robustness of LID models across diverse linguistic and real-world conditions, we selected five datasets spanning clean, noisy, informal, and speech-based text sources. These datasets vary in language coverage, domain, and text quality. Table~\ref{tab:dataset_summary} details key dataset properties.

\subsubsection{\diversmono: Monolingual data}
\textbf{FLORES-200} \citep{goyal-etal-2022-flores} provides professionally translated sentences across 200 languages, for a clean and standardized benchmark, particularly for low-resource languages.

\textbf{SMOL} \citep{caswell2025smol} is a multilingual benchmark dataset consisting of professionally translated sentence- and document-level data for 115 LRLs. SMOL highlights linguistic diversity by incorporating related, endangered, and underrepresented languages, and includes sources from diverse domains.

\textbf{TweetLID} \citep{tweetlid} is composed of annotated Twitter data, rich in abbreviations, code-switching, and informal expressions, and messy texts for evaluating robustness.

\textbf{Bloom stories} \citep{leong-etal-2022-bloom} is a dataset of publicly accessible children’s stories translated into over 100 languages, providing high-quality but simpler text that reflects natural and structured storytelling patterns that differ from other domains.

\textbf{Common Voice} \citep{commonvoice:2020} is a multilingual read-speech dataset that collects texts from language communities and records them being read aloud. For our evaluation, we focus only on the sentences.

\begin{table}[ht]
\centering
\small
\begin{minipage}[t]{0.54\textwidth}
\centering
\setlength{\tabcolsep}{2.5pt}
\begin{tabular}{llc}
\toprule
\textbf{Model} & \textbf{Technique} & \textbf{\#Langs} \\
\midrule
CLD3 & n-gram Neural Net & 100+ \\
FastText & Subword embeddings & 176 \\
Franc & Trigram frequency & 414 \\
LangDetect & Statistical + rule-based & 24 \\
LangID & Naïve Bayes & 97 \\
GlotLID & FastText  & 2000+ \\
OpenLID & FastText & 200+ \\
ConLID & FastText + SCL & 2000+ \\
\bottomrule
\end{tabular}
\caption{Evaluated LID models.}
\label{tab:lid-models}
\end{minipage}%
\hfill
\begin{minipage}[t]{0.46\textwidth}
\centering
\setlength{\tabcolsep}{2.5pt}
\begin{tabular}{lccc}
\toprule
\textbf{Dataset} & \textbf{\#Langs} & \textbf{Size} & \textbf{Domain} \\
\midrule
FLORES & 201 & 215k & Formal \\
Bloom & 101 & 1k & Stories \\
TweetLID & 8 & 17k & Informal \\
SMOL & 111 & 49k &  Diverse \\
Common Voice & 102 & 495k & Speech \\
\bottomrule
\end{tabular}
\caption{Evaluation datasets.}
\label{tab:dataset_summary}
\end{minipage}
\end{table}
\vspace{-2mm}

\subsubsection{\diverscs: CodeSwitch dataset}
\vspace{-2mm}
We identified CS corpora meeting three criteria: (i) naturally occurring intra-sentential switches, (ii) clear language-pair annotations, and (iii) minimal synthetic or weak labeling. We manually filtered these sources to retain high-quality segments, combined into a unified CS benchmark covering 9 language pairs - including both high-resource (e.g., English–Spanish) and low-resource (e.g., English-Malayalam) combinations. While prior work such as \cite{patwa-etal-2020-semeval}, \cite{aguilar2020lincecentralizedbenchmarklinguistic}, and \cite{khanuja-etal-2020-gluecos} has made notable contributions to the evaluation of code-switching, our study extends this line of research by encompassing a broader set of languages across different resource groups. In addition, we ensure dataset reliability by excluding sources of lower quality, provide a consolidated and accessible single-resource for future research, and conduct systematic benchmarking of multiple widely used LID models on the dataset.
Table~\ref{tab:cs-data-dist} summarizes the language distribution between train and test splits. 
Appendix~\ref{sec:cs_data_dist} details the characteristics of individual datasets.

\begin{table*}[ht]
\centering
\scalebox{0.8}{
\begin{tabular}{lccrr}
\toprule
\textbf{Language Pair} & \textbf{Language Code} & \textbf{Resource Class} & \textbf{\#Train} & \textbf{\#Test} \\
\midrule
English–German & en-de &5 & 30,000 & 253 \\
English–Hindi & en-hi &4 & 17,360 & 7,440 \\
Chinese–English & zh-en &5 & 15,938 & 5,940 \\
English–Tamil & en-ta & 3 & 11,237 & 4,270 \\
English–Malayalam & en-ml & 1 & 4,839 & 1,710 \\
Egyptian–English & arz-en & 3 & 2,160 & 1,812 \\
Basque–Spanish & eu-es & 4 & 1,087 & 412 \\
Arabic–English & arq-en & -- & 738 & 317 \\
English–Indonesian & en-id & 3 & 577 & 248 \\
English–Turkish & en-tr & 4 & 260 & 112 \\
\bottomrule
\end{tabular}
}
\caption{Language pair distribution in the CS dataset, with resource class on non-English language identified using \cite{joshi-etal-2020-state} framework.}
\label{tab:cs-data-dist}
\end{table*}
\vspace{-2mm}
To evaluate model performance on code-switched (CS) text, we adopt two metrics: Full Match (FM), defined as the percentage of instances where both languages in a CS sentence are correctly identified; and Partial Match (PM), defined as the percentage of instances where one of the two languages is correctly predicted but not both. Formal metric definitions are provided in Appendix~\ref{sec:formulae}.
Most LID models are trained with softmax activation, this affects their ability to extract multiple labels with a threshold, thus we select the top-2 predicted languages based on the probability scores.

\section{Results and Discussions}
\vspace{-2mm}
\begin{table*}[ht]
\centering
\footnotesize
\setlength{\tabcolsep}{4pt}

\begin{tabular}{lcccccccccccc}
\toprule
\multirow{2}{*}{\textbf{Model}} & \multicolumn{2}{c}{\textbf{Bloom}} & \multicolumn{2}{c}{\textbf{TweetLID}} & \multicolumn{2}{c}{\textbf{Common Voice}} & \multicolumn{2}{c}{\textbf{SMOL}} & \multicolumn{2}{c}{\textbf{FLORES}} & \multicolumn{2}{c}{\textbf{Avg}} \\
\cmidrule(r){2-3} \cmidrule(r){4-5} \cmidrule(r){6-7} \cmidrule(r){8-9} \cmidrule(r){10-11} \cmidrule(r){12-13} 
 & \textbf{F1}$\uparrow$ & \textbf{FPR}$\downarrow$ & \textbf{F1}$\uparrow$ & \textbf{FPR}$\downarrow$ & \textbf{F1}$\uparrow$ & \textbf{FPR}$\downarrow$ & \textbf{F1}$\uparrow$ & \textbf{FPR}$\downarrow$ & \textbf{F1}$\uparrow$ & \textbf{FPR}$\downarrow$ & \textbf{F1}$\uparrow$ & \textbf{FPR}$\downarrow$\\
\midrule
CLD3                  & 57.3 & 0.0027 & 64.4 & 0.0046 & 69.7 & 0.0022 & 15.9 & 0.0048 & 33.4 & 0.0028 & 48.2 & 0.0034 \\
Langdetect            & 52.4 & 0.0042 & 72.2 & 0.01 & 57.0 & 0.0043 & 6.3  & 0.0067 & 21.4 & 0.0037 & 41.9 & 0.0058 \\
Franc                 & 61.2 & 0.0025 & 30.5 & 0.0024 & 69.7 & 0.0012 & 36.9 & 0.0017 & 65.8 & 0.0008 & 52.8 & 0.0017 \\
LangID                & 50.7 & 0.0033 & \textbf{77.4} & 0.0041 & 69.3 & 0.0024 & 9.9  & 0.0051 & 30.0 & 0.003 & 47.4 & 0.0036 \\
FastText    & 70.5 & 0.0019 & 71.1 & 0.0026 & 72.7 & 0.0010 & 42.4 & 0.0025 & 85.6 & 0.0006 & 68.5 & 0.0017 \\
OpenLID               & 70.5 & 0.0021 & 67.2 & 0.0023 & 71.3 & 0.0011 & 46.1 & 0.0026 & 86.8 & 0.0007 & 68.4 & 0.0018\\
ConLID               & 87.2 & \textbf{0.0009}  & 59.2 & \textbf{0.0009} & \textbf{84.3} & \textbf{0.0002} & 61.6 & 0.0008 & \textbf{94.1} & \textbf{0.0002} & 77.3 & \textbf{0.0006} \\
GlotLID               & \textbf{90.3} & 0.008  & 71.8 & \textbf{0.001} & 69.3 & 0.0004 & \textbf{64.6} & \textbf{0.0007} & 93.3 & \textbf{0.0002}  & \textbf{77.8} & 0.0021 \\
\bottomrule
\end{tabular}

\caption{Full match evaluation results (Macro F1\% and False Positive Rate) across five multilingual test datasets. Higher F1 and lower FPR indicate better performance. Best results in \textbf{bold}.}
\label{tab:full-eval-results}
\end{table*}
\vspace{-2mm}

\paragraph{Large gap in performance when compared to formal domain.}
Table~\ref{tab:full-eval-results} shows models that perform well on \flores degrade sharply on informal data. ConLID, for instance, drops from 94.1 F1 on \flores to 61.6 on \smol and 59.2 F1 on \tweetlid. OpenLID and GlotLID show similar declines, losing over 30 F1 points on \smol and 20 F1 points on \tweetlid. This supports our hypothesis that non-canonical, real-world inputs remain a major challenge for LID systems.

\paragraph{Contrastive and data-rich models outperform traditional approaches, but generalization remains limited.}
ConLID and GlotLID achieve the highest average F1 scores (77.3 and 77.8 F1), outperforming traditional models like LangDetect (41.9). ConLID excels on \flores (94.1) and \commonvoice (84.3), while GlotLID leads on \bloom (90.3) and \smol (64.6), suggesting complementary strengths. However, ConLID's performance drops to 61.6 F1 on \smol and 59.2 F1 on \tweetlid, indicating it struggles to generalize despite being designed for out-of-domain and low-resource robustness. Both maintain the lowest false positive rates (FPR: 0.0006 and 0.0021), while traditional models show higher FPRs,  highlighting their reduced reliability in noisy settings.

\begin{table*}[ht]
\centering
\small
\setlength{\tabcolsep}{2pt}
\begin{tabular}{lccccccccccc}
\toprule
\multirow{2}{*}{\textbf{Model}} & \multicolumn{2}{c}{\textbf{Bloom}} & \multicolumn{2}{c}{\textbf{TweetLID}} & \multicolumn{2}{c}{\textbf{Common Voice}} & \multicolumn{2}{c}{\textbf{SMOL}} & \multicolumn{2}{c}{\textbf{FLORES}} & \textbf{Avg} \\
\cmidrule(r){2-3} \cmidrule(r){4-5} \cmidrule(r){6-7} \cmidrule(r){8-9} \cmidrule(r){10-11} \cmidrule(r){12-12} 
 & \textbf{subset} & \textbf{F1} & \textbf{subset} & \textbf{F1}& \textbf{subset} & \textbf{F1} & \textbf{subset}& \textbf{F1} & \textbf{subset} & \textbf{F1} & \textbf{F1} \\
\midrule
CLD3         & 29 & 85.4 & 6 & 64.4 & 60 & 92.7 & 18 & 84.1 & 80 & 95.5 & 84.6 \\
Langdetect   & 19 & \textbf{97.8} & 4 & 75.7 & 44 & 92.5 & 7  & 84.4 & 44 & \textbf{99.7} & \textbf{91.3} \\
Franc        & 40 & 82.4 & 6 & 30.5 & 72 & 61.9 & 46 & 76.0 & 161 & 82.9 & 69.4 \\
LangID       & 23 & 85.4 & 6 & \textbf{77.4} & 64 & 90.7 & 12 & 77.4 & 74 & 91.4 & 84.6 \\
FastText     & 30 & 90.3 & 6 & 71.1 & 81 & 70.9 & 18 & 54.7 & 99 & 93.3 & 78.5 \\
OpenLID      & 33 & 95.4 & 6 & 67.2 & 70 & 85.5 & 46 & 85.7 & 188 & 92.7 & 87.0 \\
ConLID       & 68 & 94.0 & 7 & 60.5 & 88 & \textbf{96.5} & 72 & 86.2 & 198 & 95.5 & 87.8 \\
GlotLID      & 68 & 95.4 & 6 & 71.8 & 88 & 90.9 & 72 & \textbf{87.1} & 198 & 95.7 & 89.4 \\					
\bottomrule
\end{tabular}
\caption{Partial match evaluation: F1 scores are computed on the subset of each dataset containing only languages supported by the respective model. Best results in \textbf{bold}}
\label{tab:partial-eval-results}
\end{table*}

\paragraph{Effect of language coverage.}
Table~\ref{tab:partial-eval-results} shows performance when evaluation is restricted to languages each model supports. Narrow models such as LangDetect and CLD3 improve dramatically (41.9→91.3 F1; 63.6→84.6 F1), since unsupported languages are excluded. In contrast, broad-coverage models like ConLID and GlotLID show smaller gains (+10 F1), reflecting their broad coverage. This illustrates a key trade-off: partial match inflates narrow models, while full evaluation better reflects real-world robustness. Coverage itself is therefore a critical dimension of model utility.

\vspace{-2mm}
\paragraph{LID performance drops on low-resource languages.}
Table~\ref{tab:resource-class-eval} shows performance across resource classes using the \citet{joshi-etal-2020-state} taxonomy. High-resource classes (4–5) consistently exceed 85 F1, while low-resource classes (0–2) often fall below 30 F1. Even top models like GlotLID and ConLID decline steeply in low-resource groups, underscoring that LID performance depends heavily on language resource availability.

\vspace{-2mm}
\begin{table*}[ht!]
\centering
\label{tab:fm_scores}
\begin{tabular}{lccccc}
\toprule
Label & ConLID & Openlid & Franc & Glotlid & Langdetect \\
\midrule
arz-en & 0.00 & 0.00 & 0.00 & 0.00 & 0.00 \\
zh-en  & 0.00 & \textbf{0.38} & 0.00 & 0.00 & 0.00 \\
en-id  & 0.00 & 9.27 & 0.81 & \textbf{10.08} & 0.00 \\
en-tr  & 0.00 & \textbf{4.24} & 0.00 & \textbf{4.24} & 2.54 \\
en-de  & 0.00 & 4.35 & 1.19 & \textbf{6.32} & 5.14 \\
en-hi  & \textbf{0.44} & 0.00 & 0.00 & 0.00 & 0.01 \\
eu-es  & 0.26 & \textbf{16.02} & 0.73 & 11.89 & 0.00 \\
en-ml  & \textbf{0.31} & 0.29 & 0.00 & 0.29 & 0.06 \\
en-ta  & \textbf{0.18} & 0.02 & 0.00 & 0.05 & 0.05 \\
eg-en  & 0.14 & \textbf{2.24} & 0.00 & 1.09 & 0.00 \\
\bottomrule
\end{tabular}
\caption{CodeSwitch Results: Full Match (FM) scores per language pair across models. }
\end{table*}
\vspace{-3mm}
\paragraph{LID models struggle on code-switched inputs.}
Table ~\ref{tab:fm_scores} highlights the models performance across code switch pairs in the full match (FM) setting. Models perform poorly on (FM), with most scoring near zero across all language pairs. The highest FM is OpenLID (16.0) on \texttt{eu-es}. In partial match setup, models had higher scores $\approx 48\text{–}50$, indicating that models often detect at least one language, but fail to identify both with strong bias toward dominant languages like English. ConLID, despite its design for robustness, achieves just 0.4 FM on \texttt{en-hi}. These results highlight the limitations of softmax-based LID models in handling intra-sentential multilingual inputs and emphasize the need for architectures capable of multilabel or span-based predictions. Detailed PM results in Appendix~\ref{sec:PM_codeswicting}
\vspace{-2mm}
\begin{table*}[ht!]
\centering
\small
\setlength{\tabcolsep}{1.3pt}
\begin{tabular}{lccccccccccc}
\toprule
\textbf{Dataset} & \textbf{Class} & \textbf{Langs} & \textbf{Samples} & \textbf{CLD3} & \textbf{LangDet} & \textbf{Franc} & \textbf{LangID} & \textbf{FastText} & \textbf{OpenLID} & \textbf{ConLID}  & \textbf{GlotLID} \\
\midrule
\multirow{6}{*}{\bloom} 
& 0 & 18 & 99 & 0.0 & 0.0 & 15.8 & 0.0 & 2.8 & 3.5 & 59.0 & \underline{62.5} \\
& 1 & 12 & 108 & 33.3 & 20.0 & 81.9 & 16.7 & 88.3 & 81.0 & \underline{91.2} & \underline{91.2} \\
& 2 & 4 & 61 & 79.1 & 40.0 & 88.4 & 40.0 & \textbf{99.2} & \textbf{99.2} &  \textbf{99.2} & \underline{\textbf{98.4}} \\
& 3 & 7 & 125 & 81.0 & 72.0 & 61.4 & 53.3 & 73.4 & 71.6 &  68.5 & 71.6 \\
& 4 & 5 & 78 & \underline{\textbf{100}} & \underline{\textbf{100}} & 57.7 & \underline{\textbf{100}} & 80.0 & 80.0 & 98.3 & 90.0 \\
& 5 & 4 & 359 & 97.7 & 98.2 & \textbf{91.2} & 97.6 & \underline{99.0} & 97.1 & 96.5 & 97.2\\
\midrule
\multirow{6}{*}{Com. Voice} 
& 0 & 3  & 3463   & 0.0  & 0.0  & 31.9 & 0.0  & 0.0  & 0.0  & \underline{\textbf{100}} & \underline{\textbf{100}} \\
& 1 & 30 & 109697 & 29.5 & 10.3 & 47.5 & 31.1 & 65.2 & 59.2 & \underline{88.3} & 88.2 \\
& 2 & 7  & 5490   & 68.6 & 33.1 & 60.3 & 53.6 & 84.2 & \underline{84.8} & 83.8  & 84.7 \\
& 3 & 20 & 104409 & \underline{90.7} & 69.9 & \textbf{61.1} & 85.1 & \textbf{87.2} & \textbf{87.0} & 80.1 & 83.4 \\
& 4 & 17 & 122395 & 91.0 & 83.6 & 48.7 & 91.1 & 79.0 & 77.8 & 94.1 & \underline{98.9} \\
& 5 & 7 & 100695 & \textbf{93.9} & \textbf{90.6} & 53.7 & \underline{\textbf{96.4}} & 71.9 & 78.5 & 68.6 & 81.8 \\  
\midrule
\multirow{5}{*}{\smol} 
& 0 & 22 & 14109 & 0.0  & 0.0  & 26.8 & 0.0  & 17.5 & 15.6 & 54.5 & \underline{55.3} \\
& 1 & 28 & 24303 & 14.2 & 3.7  & 40.6 & 9.0  & 65.3 & 61.9 & 68.1 & \underline{68.3} \\
& 2 & 10 & 9493  & 38.8 & 0.0  & \textbf{67.5} & 21.2 & \textbf{88.9} & \underline{\textbf{89.4}} & \textbf{83.6} & \textbf{85.9} \\
& 3 & 2  & 1726  & 49.2 & 49.6 & 47.3 & 39.6 & 49.9 & \underline{78.7} & 76.5 & 72.7 \\
& 5 & 2  & 1726  & \textbf{99.2} & \underline{\textbf{99.5}} & 37.1 & \textbf{99.1} & 49.8 & 49.2 & 49.5 & 50.0 \\
\midrule
\multirow{6}{*}{\flores} 
& 0 & 13 & 13156 & 7.6  & 0.0  & 52.8 & 7.7  & 74.7 & \underline{74.2} & 87.8 & 87.1 \\
& 1 & 70 & 75900 & 31.1 & 10.1 & 64.8 & 23.3 & 91.8 & 92.0 & \underline{\textbf{98.7}} & 98.4 \\
& 2 & 17 & 18216 & 64.5 & 11.8 & 86.1 & 52.9 & \textbf{99.6} & \underline{\textbf{99.8}} & 98.5 & 99.2 \\
& 3 & 22 & 22264 & 87.2 & 67.8 & 84.3 & 84.4 & 92.9 & \underline{97.2} & 94.2 & 96.0 \\
& 4 & 17 & 18216 & \textbf{97.1} & \textbf{87.9} & 83.2 & \textbf{99.0} & 99.3 & 98.7 & 97.6 & \underline{99.6} \\
& 5 & 6  & 7084  & 83.0 & 83.2 & \textbf{95.4} & 83.3 & 83.0 & 83.0 & 98.3 & \underline{\textbf{99.7}} \\
\midrule
\multirow{3}{*}{Tweet} 
& 3 & 1 & 415    & 68.9 & 0.0  & \textbf{41.9} & 47.2 & \underline{\textbf{72.8}} & \underline{\textbf{72.8}} & \textbf{71.3} & 65.8 \\
& 4 & 3 & 3643  & \textbf{69.7} & 55.1 & 28.4 & 68.5 & 53.1 & 56.3 & 67.8 & \underline{\textbf{80.4}} \\
& 5 & 2 & 12448 & 58.2 & \textbf{74.5} & 30.6 & \underline{\textbf{78.8}} & 71.1 & 72.0 & 51.2 & 75.4 \\
\bottomrule
\end{tabular}
\caption{Model performance across \citet{joshi-etal-2020-state} resource classes  across five multilingual
test datasets. Each row reports the F1 score averaged over languages belonging to a resource class. The best score per model per dataset in \textbf{bold}, and the best score per resource class per dataset is \underline{underlined}}
\label{tab:resource-class-eval}
\end{table*}
\vspace{-6mm}

\section{Conclusion}
\vspace{-3mm}
In this paper, we evaluated LID models in diverse domains and code-switched settings. Recent LID models perform well on clean text; our findings show they struggle with noisy, informal, and mixed-language input. This confirms that the findings of \cite{lid_web} and the issues identified by \cite{lid_wild} continue to persist. Code-switching has shown to be a challenging setup and is probably the most realistic for low-resource languages, yet LID systems display an inability to handle code-switching. 
Looking forward, we recommend training on data that spans multiple domains, including realistic and noisy text, to improve generalization. We also hope this work encourages evaluation beyond clean benchmarks and drives the development of more robust LID models that better reflect real-world multilingual use.

\section{Acknowledgement}
DIA acknowledges the support of the Natural Sciences and Engineering Research Council of Canada (NSERC)---Discovery Grant program, as well as funding from IVADO and the Canada First Research
Excellence Fund.  

\section{Limitations}
Our study evaluates language identification at the sentence and paragraph level. We do not explore word-level LID, which is especially relevant in code-switched settings and comes with its own set of challenges.

We did not evaluate on large language models (LLMs) or transformer-based approaches. Although these models are increasingly capable of multilingual tasks, they are not typically optimized or deployed for lightweight, real-time LID, and often require substantial compute resources. Our focus is instead on systems explicitly designed for LID that are representative of current practice and feasible in real-world applications. We encourage future work to investigate how LLMs and transformer-based models can complement or extend dedicated LID systems, particularly for low-resource and code-switched scenarios. 

Our evaluation relies on multiple datasets with differing label formats, some provide language names, others provide codes. To ensure consistency, we use the \texttt{pycountry} library to map all labels to ISO codes and scripts, and adapt them to the requirements of each model (e.g., CLD3 outputs ISO 639-1 codes).\footnote{https://github.com/pycountry/pycountry}


\bibliography{colm2025_conference}
\bibliographystyle{colm2025_conference}

\clearpage
\newpage
\appendix
\clearpage

\section{Code-Switching Dataset Distribution}
Table~\ref{tab:cs-datasetsources} details the individual datasets collected and combined into \csbench, for each dataset, when original train-test splits were available, we preserved them; otherwise, a 70\%/30\% split was used. 
\label{sec:cs_data_dist}
\begin{table*}[ht]
\centering
\setlength{\tabcolsep}{2.3pt}
\begin{tabular}{@{}l l c c l@{}}
\toprule
\textbf{Dataset Source} & \textbf{Languages} & \textbf{Total Size} & \textbf{Test Size} & \textbf{Reference} \\
\midrule
SemEval-2020 Task 9 & \makecell[l]{English–Hindi \\ English–Spanish} & 38,813 & 6,000 & \citet{patwa-etal-2020-semeval} \\
\midrule
TongueSwitcher & English–German & 30,253 & 253 & \citet{sterner2023tongueswitcher} \\
\midrule
Dravidian-CodeMix & \makecell[l]{English–Malayalam \\ English–Tamil} & 15,664 & 5,980 & \citet{chakravarthi-etal-2020-sentiment,chakravarthi-etal-2020-corpus} \\
\midrule
CroCoSum & Chinese–English & 12,989 & 5,567 & \citet{zhang-eickhoff-2024-crocosum} \\
\midrule
ASCEND & Chinese–English & 3,320 & 373 & \citet{lovenia2022ascend} \\
\midrule
HinGE & English–Hindi & 4,799 & 1,440 & \citet{srivastava2021hingedatasetgenerationevaluation} \\
\midrule
ArzEn-ST & English–Egyptian & 6,284 & 1,812 & \citet{hamed-etal-2022-arzen} \\
\midrule
BaSCo-Corpus & Basque–Spanish & 1,372 & 412 & \citet{basco} \\
\midrule
Barik et al. & English–Indonesian & 825 & 248 & \citet{barik-etal-2019-normalization} \\
\midrule
Sabty (2021) & Arabic–English & 1,055 & 317 & \citet{sabty2021language} \\
\midrule
Yirmibesoglu \& Eryigit & English–Turkish & 391 & 112 & \citet{yirmibesoglu-eryigit-2018-detecting} \\
\bottomrule
\end{tabular}
\caption{Summary of datasets used, including language coverage, total sample size, and references.}
\label{tab:cs-datasetsources}
\end{table*}

\section{Models Evaluated}
\label{sec:model_descriptions}
\paragraph{Google's Compact Language Detector 3 (CLD3)} an n-gram neural network model optimized for language identification.~\footnote{\url{https://github.com/google/cld3}} We use the python implementation. ~\footnote{\url{https://pypi.org/project/gcld3/}}
\vspace{-3mm}
\paragraph{FastText LID} a FastText-based language identification model trained primarily on Wikipedia text, covering 176 languages.~\citep{joulin2016fasttext}
\vspace{-3mm}
\paragraph{Franc} a trigram frequency-based detector optimized for efficiency covering 414 languages.~\footnote{\url{https://github.com/wooorm/franc/tree/main/packages/franc-all}} We use the python implementation.~\footnote{\url{https://github.com/cyb3rk0tik/pyfranc?tab=readme-ov-file}}. 
\vspace{-3mm}
\paragraph{LangDetect} a statistical and rule-based detector adapted from Google's language-detection library covering 24 targeted languages.\footnote{\url{https://github.com/fedelopez77/langdetect?tab=readme-ov-file}} 
\vspace{-3mm}
\paragraph{LangID.py} a standalone language identification tool trained on 97 languages~\citep{lui-baldwin-2012-langid}\vspace{-3mm}
\paragraph{GlotLID} a multilingual LID model trained on a mixture of religious texts, Wikipedia, and web data, covering over 2000 languages.~\citep{kargaran-etal-2023-glotlid}
\vspace{-3mm}
\paragraph{OpenLID} a FastText-based opensource LID model trained on more than 200 languages.~\citep{burchell-etal-2024-code}
\vspace{-3mm}
\paragraph{ConLID} a recent FastText-based LID that introduces training with  supervised contrastive learning (SCL) approach to improve LID robustness to out-of-domain data for low-resource languages.~\citep{conlid}

\clearpage

\section{Partial Match - Codeswitching}
\label{sec:PM_codeswicting}
In this setup we evaluate the model's ability to detect at least one language in the codeswitch pair. Table~\ref{tab:pm_mfl_updated} reports the PM score (formular in Appendix~\ref{sec:formulae} and the most frequently predicted language (MFL).
\begin{table*}[ht]
\centering
\resizebox{0.99\textwidth}{!}{%
\begin{tabular}{lcccccccccccccc}
\toprule
\textbf{Label} 
& \multicolumn{2}{c}{\textbf{ConLID}} 
& \multicolumn{2}{c}{\textbf{Openlid}} 
& \multicolumn{2}{c}{\textbf{Franc}} 
& \multicolumn{2}{c}{\textbf{Glotlid}} 
& \multicolumn{2}{c}{\textbf{Langdetect}} \\
\cmidrule(lr){2-3} \cmidrule(lr){4-5} \cmidrule(lr){6-7} \cmidrule(lr){8-9} \cmidrule(lr){10-11} \cmidrule(lr){12-13}
 & PM $\uparrow$ & MFL & PM $\uparrow$ & MFL & PM $\uparrow$ & MFL & PM $\uparrow$ & MFL & PM $\uparrow$ & MFL \\
\midrule
zh-en  & 48.4 & en & 50.0 & zh & 49.8 & en & 49.7 & en & 49.7 & en \\
en-id  & 50.0 & en & 49.7 & id & 50.0 & id & 49.8 & id & 49.4 & en \\
en-tr  & 48.9 & en & 49.7 & tr & 50.0 & tr & 49.8 & tr & 50.0 & tr \\
en-de  & 48.9 & en & 45.4 & de & 50.0 & de & 50.0 & de & 50.0 & de \\
arz-en & 50.0 & en & 49.0 & en & 50.0 & en & 49.8 & en & 49.9 & en \\
en-hi  & 49.5 & hi & 49.2 & en & 50.0 & en & 49.6 & en & 49.9 & en \\
eu-es  & 50.0 & eu & 49.4 & eu & 49.6 & eu & 49.6 & eu & 49.8 & es \\
en-ml  & 45.2 & ml & 47.9 & ml & 50.0 & ml & 47.8 & ml & 49.3 & en \\
en-ta  & 44.3 & en & 48.7 & en & 50.0 & en & 48.9 & en & 49.4 & en \\
en-eg  & 48.3 & en & 46.0 & eg & 50.0 & en & 46.7 & eg & 45.3 & en \\
\bottomrule
\end{tabular}%
}
\caption{Partial Match (in percentages) (PM $\uparrow$) and Most Frequent Language (MFL) for each language pair. Bolded PM values indicate the best model for that row. }
\label{tab:pm_mfl_updated}
\end{table*}

\section{Challenges with collecting CS Datasets}
\label{sec: cs-data-challenges}

\begin{itemize}
    \item \textbf{Lack of Accessibility:} Many prior CS datasets are not publicly available, limiting reproducibility and comparative studies.
    \item \textbf{Mixed Datasets:} Available resources often contain single-label and code-switched sentences mixed together, which increases the difficulty of automatic or manual filtering which might not always be possible
    \item \textbf{Synthetic Data:} Some datasets rely on artificially generated CS text, which often lacks the natural flow of codeswitching.
    \item \textbf{Inadequate Labeling:} Several datasets mostly the dialect datasets such as \citet{hamed2024zaebucspokenmultilingualmultidialectalarabicenglish} classified the CS labels as broad categories such as \textit{"MSA-dialectal code-switching"} and \textit{"imperfect MSA"} rather than having labels specifying the dialects in the CS sentence.
\end{itemize}

\section{Formulae used in Evaluation}
\label{sec:formulae}

\[
\text{False Positive Rate (FPR)} = \frac{\text{False Positives (FP)}}{\text{False Positives (FP)} + \text{True Negatives (TN)}}
\]

\[
\text{Full Match Accuracy}(\ell) = \frac{\sum_{i=1}^{N_\ell} \mathbf{1}(\hat{y}_i = y_i)}{N_\ell}
\]

\[
\text{Partial Match}(\ell) = \frac{1}{N_\ell} \sum_{i=1}^{N_\ell} \frac{|S_i^\text{true} \cap S_i^\text{pred}|}{|S_i^\text{true} \cup S_i^\text{pred}|}
\quad \text{where} \quad S_i^\text{true} \neq S_i^\text{pred}
\quad \text{and} \quad |S_i^\text{true} \cap S_i^\text{pred}| > 0
\]

\section{Languages Evaluated}
\label{sec:language_results}
Here we detail the languages evaluated, their presence in each dataset and each model F1 scores
\insertlanguagedetail


\end{document}

%% file: full_language_eval_table_reordered.tex
\newcommand{\insertlanguagedetail}{\begin{table}[htbp]
\centering
\small
\caption{Full Language Evaluation Summary}
\label{tab:lang_eval_full}
\setlength{\tabcolsep}{1.8pt}
\begin{tabular}{llrp{0.8cm}lllllrrrrrrrr}
\toprule
Code & Name & Size & Class & Blm & Flr & TLD & Sml & CV & Open & CLD3 & ConL & FastT & Franc & Glot & LD & Lid \\
\midrule
aar & Afar & 6904 & 0.0 & \ding{55} & \ding{55} & \ding{55} & \ding{51} & \ding{55} & 0.0 & 0.0 & 0.0 & 0.0 & \underline{0.84} & 0.0 & 0.0 & 0.0 \\
abk & Abkhazian & 72992 & 1.0 & \ding{55} & \ding{55} & \ding{55} & \ding{55} & \ding{51} & 0.0 & 0.0 & \underline{1.0} & 0.96 & \underline{0.9} & \underline{1.0} & 0.0 & 0.0 \\
ace & Achinese & 16192 & - & \ding{55} & \ding{51} & \ding{55} & \ding{55} & \ding{55} & \underline{0.98} & 0.0 & \underline{1.0} & 0.98 & \underline{0.48} & \underline{0.99} & 0.0 & 0.0 \\
ach & Acoli & 6904 & - & \ding{55} & \ding{55} & \ding{55} & \ding{51} & \ding{55} & 0.0 & 0.0 & \underline{0.99} & 0.0 & 0.0 & \underline{0.97} & 0.0 & 0.0 \\
acm & Mesopotamian Arabic & 8096 & - & \ding{55} & \ding{51} & \ding{55} & \ding{55} & \ding{55} & \underline{0.02} & 0.0 & \underline{0.22} & 0.0 & 0.0 & \underline{0.4} & 0.0 & 0.0 \\
acq & Ta'izzi-Adeni Arabic & 8096 & - & \ding{55} & \ding{51} & \ding{55} & \ding{55} & \ding{55} & \underline{0.0} & 0.0 & 0.0 & 0.0 & 0.0 & 0.0 & 0.0 & 0.0 \\
aeb & Tunisian Arabic & 14992 & - & \ding{55} & \ding{51} & \ding{55} & \ding{51} & \ding{55} & \underline{0.38} & 0.0 & \underline{0.63} & 0.0 & 0.0 & \underline{0.66} & 0.0 & 0.0 \\
\end{tabular}
\end{table}

\begin{table}[htbp]
\centering
\small
\label{tab:lang_eval_full}
\setlength{\tabcolsep}{1.8pt}
\begin{tabular}{llrp{0.8cm} lllll rrrrrrrr}
\toprule
Code & Name & Size & Class & Blm & Flr & TLD & Sml & CV & Open & CLD3 & ConL & FastT & Franc & Glot & LD & Lid \\
\midrule

bwx & Bu-Nao Bunu & 8 & - & \ding{51} & \ding{55} & \ding{55} & \ding{55} & \ding{55} & 0.0 & 0.0 & 0.0 & 0.0 & 0.0 & 0.0 & 0.0 & 0.0 \\
bzi & Bisu & 48 & 0.0 & \ding{51} & \ding{55} & \ding{55} & \ding{55} & \ding{55} & 0.0 & 0.0 & \underline{1.0} & 0.0 & 0.0 & \underline{1.0} & 0.0 & 0.0 \\
cak & Kaqchikel & 48 & - & \ding{51} & \ding{55} & \ding{55} & \ding{55} & \ding{55} & 0.0 & 0.0 & \underline{1.0} & 0.0 & \underline{1.0} & \underline{1.0} & 0.0 & 0.0 \\
cat & Catalan & 158328 & 4.0 & \ding{55} & \ding{51} & \ding{51} & \ding{55} & \ding{51} & \underline{0.33} & \underline{0.89} & \underline{0.9} & \underline{0.33} & \underline{0.26} & \underline{0.33} & \underline{0.93} & \underline{0.87} \\
cbr & Cashibo-Cacataibo & 8 & - & \ding{51} & \ding{55} & \ding{55} & \ding{55} & \ding{55} & 0.0 & 0.0 & \underline{1.0} & 0.0 & \underline{1.0} & \underline{1.0} & 0.0 & 0.0 \\
ccp & Chakma & 3656 & - & \ding{55} & \ding{55} & \ding{55} & \ding{51} & \ding{55} & 0.0 & 0.0 & \underline{0.43} & 0.0 & 0.0 & \underline{0.63} & 0.0 & 0.0 \\
ceb & Cebuano & 8424 & 3.0 & \ding{51} & \ding{51} & \ding{55} & \ding{55} & \ding{55} & \underline{0.96} & \underline{0.93} & \underline{0.87} & \underline{0.98} & \underline{0.7} & \underline{0.9} & 0.0 & 0.0 \\
\end{tabular}
\end{table}

\begin{table}[htbp]
\centering
\small
\label{tab:lang_eval_full}
\setlength{\tabcolsep}{1.8pt}
\begin{tabular}{llrp{0.8cm}lllllrrrrrrrr}
\toprule
Code & Name & Size & Class & Blm & Flr & TLD & Sml & CV & Open & CLD3 & ConL & FastT & Franc & Glot & LD & Lid \\
\midrule

hat & Haitian & 8304 & 2.0 & \ding{51} & \ding{51} & \ding{55} & \ding{55} & \ding{55} & \underline{1.0} & \underline{0.5} & \underline{1.0} & \underline{1.0} & \underline{0.99} & \underline{0.98} & 0.0 & \underline{0.49} \\
hau & Hausa & 20400 & 2.0 & \ding{51} & \ding{51} & \ding{55} & \ding{51} & \ding{51} & \underline{0.99} & \underline{0.98} & \underline{0.99} & 0.98 & \underline{0.94} & \underline{0.99} & 0.0 & 0.0 \\
hbb & Huba & 16 & - & \ding{51} & \ding{55} & \ding{55} & \ding{55} & \ding{55} & 0.0 & 0.0 & 0.0 & 0.0 & 0.0 & 0.0 & 0.0 & 0.0 \\
heb & Hebrew & 8096 & 3.0 & \ding{55} & \ding{51} & \ding{55} & \ding{55} & \ding{55} & \underline{1.0} & \underline{0.0} & \underline{1.0} & \underline{1.0} & \underline{1.0} & \underline{1.0} & \underline{1.0} & \underline{1.0} \\
hin & Hindi & 31808 & 4.0 & \ding{51} & \ding{51} & \ding{55} & \ding{55} & \ding{51} & \underline{0.97} & \underline{0.97} & \underline{0.98} & \underline{1.0} & \underline{0.52} & \underline{0.83} & \underline{0.99} & \underline{0.96} \\
hne & Chhattisgarhi & 8096 & - & \ding{55} & \ding{51} & \ding{55} & \ding{55} & \ding{55} & \underline{0.93} & 0.0 & \underline{0.97} & 0.89 & 0.0 & \underline{0.95} & 0.0 & 0.0 \\
hoc & Ho & 3200 & 0.0 & \ding{55} & \ding{55} & \ding{55} & \ding{51} & \ding{55} & 0.0 & 0.0 & \underline{1.0} & 0.0 & 0.0 & \underline{0.93} & 0.0 & 0.0 \\
\end{tabular}
\end{table}

\begin{table}[htbp]
\centering
\small
\label{tab:lang_eval_full}
\setlength{\tabcolsep}{1.8pt}
\begin{tabular}{llrp{0.8cm} lllll rrrrrrrr}
\toprule
Code & Name & Size & Class & Blm & Flr & TLD & Sml & CV & Open & CLD3 & ConL & FastT & Franc & Glot & LD & Lid \\
\midrule
lao & Lao & 7118 & 2.0 & \ding{55} & \ding{51} & \ding{55} & \ding{55} & \ding{55} & \underline{1.0} & \underline{0.96} & \underline{1.0} & \underline{1.0} & \underline{1.0} & \underline{1.0} & 0.0 & \underline{1.0} \\
lav & Latvian & 18104 & 3.0 & \ding{55} & \ding{55} & \ding{55} & \ding{55} & \ding{51} & 0.0 & \underline{0.82} & 0.0 & \underline{0.0} & 0.0 & 0.0 & \underline{0.93} & \underline{0.87} \\
lbr & Lohorung & 8 & - & \ding{51} & \ding{55} & \ding{55} & \ding{55} & \ding{55} & 0.0 & 0.0 & 0.0 & 0.0 & 0.0 & 0.0 & 0.0 & 0.0 \\
lep & Lepcha & 3199 & 0.0 & \ding{55} & \ding{55} & \ding{55} & \ding{51} & \ding{55} & 0.0 & 0.0 & 0.0 & 0.0 & 0.0 & 0.0 & 0.0 & 0.0 \\
lhu & Lahu & 32 & 0.0 & \ding{51} & \ding{55} & \ding{55} & \ding{55} & \ding{55} & 0.0 & 0.0 & \underline{1.0} & 0.0 & 0.0 & \underline{1.0} & 0.0 & 0.0 \\
lif & Limbu & 3201 & 0.0 & \ding{55} & \ding{55} & \ding{55} & \ding{51} & \ding{55} & 0.0 & 0.0 & \underline{1.0} & 0.0 & 0.0 & \underline{1.0} & 0.0 & 0.0 \\
lij & Ligurian & 8096 & 1.0 & \ding{55} & \ding{51} & \ding{55} & \ding{55} & \ding{55} & \underline{1.0} & 0.0 & \underline{1.0} & 0.96 & \underline{0.91} & \underline{1.0} & 0.0 & 0.0 \\
\end{tabular}
\end{table}

\begin{table}[htbp]
\centering
\small
\caption{Full Language Evaluation Summary}
\label{tab:lang_eval_full}
\setlength{\tabcolsep}{1.8pt}
\begin{tabular}{llrp{0.8cm} lllll rrrrrrrr}
\toprule
Code & Name & Size & Class & Blm & Flr & TLD & Sml & CV & Open & CLD3 & ConL & FastT & Franc & Glot & LD & Lid \\
\midrule
oci & Occitan (post 1500) & 16328 & - & \ding{55} & \ding{51} & \ding{55} & \ding{55} & \ding{51} & \underline{0.91} & 0.0 & \underline{0.97} & \underline{0.94} & \underline{0.64} & \underline{0.94} & 0.0 & \underline{0.49} \\
omw & South Tairora & 8 & - & \ding{51} & \ding{55} & \ding{55} & \ding{55} & \ding{55} & 0.0 & 0.0 & \underline{1.0} & 0.0 & 0.0 & \underline{1.0} & 0.0 & 0.0 \\
ori & Oriya (macrolanguage) & 1752 & - & \ding{55} & \ding{55} & \ding{55} & \ding{55} & \ding{51} & 0.98 & 0.0 & 0.0 & \underline{0.96} & 0.0 & 0.98 & 0.0 & \underline{1.0} \\
orm & Oromo & 6904 & 1.0 & \ding{55} & \ding{55} & \ding{55} & \ding{51} & \ding{55} & 0.0 & 0.0 & 0.0 & 0.0 & 0.0 & 0.0 & 0.0 & 0.0 \\
ory & Odia & 8028 & 1.0 & \ding{55} & \ding{51} & \ding{55} & \ding{55} & \ding{55} & \underline{1.0} & 0.0 & \underline{1.0} & 1.0 & 0.0 & \underline{1.0} & 0.0 & 0.0 \\
pag & Pangasinan & 8096 & 1.0 & \ding{55} & \ding{51} & \ding{55} & \ding{55} & \ding{55} & \underline{1.0} & 0.0 & \underline{1.0} & 1.0 & 0.0 & \underline{1.0} & 0.0 & 0.0 \\
pan & Panjabi & 18312 & - & \ding{55} & \ding{51} & \ding{55} & \ding{51} & \ding{51} & \underline{0.33} & \underline{0.66} & \underline{0.66} & \underline{0.33} & \underline{0.33} & \underline{0.33} & \underline{0.67} & \underline{0.67} \\
\end{tabular}
\end{table}

\begin{table}[htbp]
\centering
\small
\caption{Full Language Evaluation Summary}
\label{tab:lang_eval_full}
\setlength{\tabcolsep}{1.8pt}
\begin{tabular}{llrp{0.8cm} lllll rrrrrrrr}
\toprule
Code & Name & Size & Class & Blm & Flr & TLD & Sml & CV & Open & CLD3 & ConL & FastT & Franc & Glot & LD & Lid \\
\midrule
tet & Tetum & 8 & 1.0 & \ding{51} & \ding{55} & \ding{55} & \ding{55} & \ding{55} & 0.0 & 0.0 & \underline{0.0} & 0.0 & \underline{1.0} & \underline{0.0} & 0.0 & 0.0 \\
tgk & Tajik & 8112 & 1.0 & \ding{51} & \ding{51} & \ding{55} & \ding{55} & \ding{55} & \underline{1.0} & \underline{1.0} & \underline{1.0} & \underline{1.0} & \underline{1.0} & \underline{1.0} & 0.0 & 0.0 \\
tha & Thai & 96032 & 3.0 & \ding{51} & \ding{51} & \ding{55} & \ding{55} & \ding{51} & \underline{1.0} & \underline{0.99} & \underline{1.0} & \underline{1.0} & \underline{0.99} & \underline{1.0} & \underline{1.0} & \underline{1.0} \\
the & Chitwania Tharu & 8 & - & \ding{51} & \ding{55} & \ding{55} & \ding{55} & \ding{55} & 0.0 & 0.0 & 0.0 & 0.0 & 0.0 & 0.0 & 0.0 & 0.0 \\
thl & Dangaura Tharu & 144 & - & \ding{51} & \ding{55} & \ding{55} & \ding{55} & \ding{55} & 0.0 & 0.0 & \underline{0.94} & 0.0 & 0.0 & \underline{1.0} & 0.0 & 0.0 \\
tig & Tigre & 7 & - & \ding{55} & \ding{55} & \ding{55} & \ding{55} & \ding{51} & 0.0 & 0.0 & \underline{1.0} & 0.0 & 0.0 & \underline{0.0} & - & 0.0 \\
tir & Tigrinya & 13148 & 2.0 & \ding{55} & \ding{51} & \ding{55} & \ding{51} & \ding{51} & \underline{1.0} & 0.0 & \underline{0.56} & 1.0 & \underline{0.77} & \underline{0.62} & 0.0 & 0.0 \\
\bottomrule
\end{tabular}
\caption{Language details: This table presents the languages evaluated in this project, including total sample size across datasets and the avg model performance per language. Resource class is based on \citet{joshi-etal-2020-state}. We abbreviate the dataset names \bloom \flores \tweetlid \smol \commonvoice as Blm, Flr, TLD, SML and CV. Similarly we abbreviate model names OpenLID, ConLID, FastText, Franc, GlotLID, LangDetect and LangID as Open, ConL, FastT, Glot, LD and Lid. We \underline{underline} model scores for languages officially supported by the model.}
\end{table}

}

%% file: colm2025_conference.bbl
\begin{thebibliography}{44}
\providecommand{\natexlab}[1]{#1}
\providecommand{\url}[1]{\texttt{#1}}
\expandafter\ifx\csname urlstyle\endcsname\relax
  \providecommand{\doi}[1]{doi: #1}\else
  \providecommand{\doi}{doi: \begingroup \urlstyle{rm}\Url}\fi

\bibitem[Adebara et~al.(2022)Adebara, Elmadany, Abdul-Mageed, and Inciarte]{adebara-etal-2022-afrolid}
Ife Adebara, AbdelRahim Elmadany, Muhammad Abdul-Mageed, and Alcides Inciarte.
\newblock {A}fro{LID}: A neural language identification tool for {A}frican languages.
\newblock In Yoav Goldberg, Zornitsa Kozareva, and Yue Zhang (eds.), \emph{Proceedings of the 2022 Conference on Empirical Methods in Natural Language Processing}, pp.\  1958--1981, Abu Dhabi, United Arab Emirates, December 2022. Association for Computational Linguistics.
\newblock \doi{10.18653/v1/2022.emnlp-main.128}.
\newblock URL \url{https://aclanthology.org/2022.emnlp-main.128/}.

\bibitem[Aguilar et~al.(2020)Aguilar, Kar, and Solorio]{aguilar2020lincecentralizedbenchmarklinguistic}
Gustavo Aguilar, Sudipta Kar, and Thamar Solorio.
\newblock Lince: A centralized benchmark for linguistic code-switching evaluation, 2020.
\newblock URL \url{https://arxiv.org/abs/2005.04322}.

\bibitem[Aguirre et~al.(2022)Aguirre, Garc{\'i}a-Sardi{\~n}a, Serras, M{\'e}ndez, and L{\'o}pez]{basco}
Maia Aguirre, Laura Garc{\'i}a-Sardi{\~n}a, Manex Serras, Ariane M{\'e}ndez, and Jacobo L{\'o}pez.
\newblock {B}a{SC}o: An annotated {B}asque-{S}panish code-switching corpus for natural language understanding.
\newblock In Nicoletta Calzolari, Fr{\'e}d{\'e}ric B{\'e}chet, Philippe Blache, Khalid Choukri, Christopher Cieri, Thierry Declerck, Sara Goggi, Hitoshi Isahara, Bente Maegaard, Joseph Mariani, H{\'e}l{\`e}ne Mazo, Jan Odijk, and Stelios Piperidis (eds.), \emph{Proceedings of the Thirteenth Language Resources and Evaluation Conference}, pp.\  3158--3163, Marseille, France, June 2022. European Language Resources Association.
\newblock URL \url{https://aclanthology.org/2022.lrec-1.338/}.

\bibitem[Ardila et~al.(2020)Ardila, Branson, Davis, Henretty, Kohler, Meyer, Morais, Saunders, Tyers, and Weber]{commonvoice:2020}
R.~Ardila, M.~Branson, K.~Davis, M.~Henretty, M.~Kohler, J.~Meyer, R.~Morais, L.~Saunders, F.~M. Tyers, and G.~Weber.
\newblock Common voice: A massively-multilingual speech corpus.
\newblock In \emph{Proceedings of the 12th Conference on Language Resources and Evaluation (LREC 2020)}, pp.\  4211--4215, 2020.

\bibitem[Baldwin \& Lui(2010)Baldwin and Lui]{baldwin-lui-2010-language}
Timothy Baldwin and Marco Lui.
\newblock Language identification: The long and the short of the matter.
\newblock In Ron Kaplan, Jill Burstein, Mary Harper, and Gerald Penn (eds.), \emph{Human Language Technologies: The 2010 Annual Conference of the North {A}merican Chapter of the Association for Computational Linguistics}, pp.\  229--237, Los Angeles, California, June 2010. Association for Computational Linguistics.
\newblock URL \url{https://aclanthology.org/N10-1027/}.

\bibitem[Barik et~al.(2019)Barik, Mahendra, and Adriani]{barik-etal-2019-normalization}
Anab~Maulana Barik, Rahmad Mahendra, and Mirna Adriani.
\newblock Normalization of {I}ndonesian-{E}nglish code-mixed {T}witter data.
\newblock In \emph{Proceedings of the 5th Workshop on Noisy User-generated Text (W-NUT 2019)}, pp.\  417--424, Hong Kong, China, November 2019. Association for Computational Linguistics.
\newblock \doi{10.18653/v1/D19-5554}.
\newblock URL \url{https://aclanthology.org/D19-5554}.

\bibitem[Burchell et~al.(2023)Burchell, Birch, Bogoychev, and Heafield]{burchell-etal-2023-open}
Laurie Burchell, Alexandra Birch, Nikolay Bogoychev, and Kenneth Heafield.
\newblock An open dataset and model for language identification.
\newblock In Anna Rogers, Jordan Boyd-Graber, and Naoaki Okazaki (eds.), \emph{Proceedings of the 61st Annual Meeting of the Association for Computational Linguistics (Volume 2: Short Papers)}, pp.\  865--879, Toronto, Canada, July 2023. Association for Computational Linguistics.
\newblock \doi{10.18653/v1/2023.acl-short.75}.
\newblock URL \url{https://aclanthology.org/2023.acl-short.75/}.

\bibitem[Burchell et~al.(2024)Burchell, Birch, Thompson, and Heafield]{burchell-etal-2024-code}
Laurie Burchell, Alexandra Birch, Robert Thompson, and Kenneth Heafield.
\newblock Code-switched language identification is harder than you think.
\newblock In Yvette Graham and Matthew Purver (eds.), \emph{Proceedings of the 18th Conference of the European Chapter of the Association for Computational Linguistics (Volume 1: Long Papers)}, pp.\  646--658, St. Julian{'}s, Malta, March 2024. Association for Computational Linguistics.
\newblock URL \url{https://aclanthology.org/2024.eacl-long.38/}.

\bibitem[Carter et~al.(2013)Carter, Weerkamp, and Tsagkias]{10.1007/s10579-012-9195-y}
Simon Carter, Wouter Weerkamp, and Manos Tsagkias.
\newblock Microblog language identification: overcoming the limitations of short, unedited and idiomatic text.
\newblock \emph{Lang. Resour. Eval.}, 47\penalty0 (1):\penalty0 195–215, March 2013.
\newblock ISSN 1574-020X.
\newblock \doi{10.1007/s10579-012-9195-y}.
\newblock URL \url{https://doi.org/10.1007/s10579-012-9195-y}.

\bibitem[Caswell et~al.(2020)Caswell, Breiner, van Esch, and Bapna]{lid_wild}
Isaac Caswell, Theresa Breiner, Daan van Esch, and Ankur Bapna.
\newblock Language {ID} in the wild: Unexpected challenges on the path to a thousand-language web text corpus.
\newblock In Donia Scott, Nuria Bel, and Chengqing Zong (eds.), \emph{Proceedings of the 28th International Conference on Computational Linguistics}, pp.\  6588--6608, Barcelona, Spain (Online), December 2020. International Committee on Computational Linguistics.
\newblock \doi{10.18653/v1/2020.coling-main.579}.
\newblock URL \url{https://aclanthology.org/2020.coling-main.579/}.

\bibitem[Caswell et~al.(2025)Caswell, Nielsen, Luo, Cherry, Kovacs, Shemtov, Talukdar, Tewari, Diane, Doumbouya, Diane, and Cissé]{caswell2025smol}
Isaac Caswell, Elizabeth Nielsen, Jiaming Luo, Colin Cherry, Geza Kovacs, Hadar Shemtov, Partha Talukdar, Dinesh Tewari, Baba~Mamadi Diane, Koulako~Moussa Doumbouya, Djibrila Diane, and Solo~Farabado Cissé.
\newblock Smol: Professionally translated parallel data for 115 under-represented languages, 2025.
\newblock URL \url{https://arxiv.org/abs/2502.12301}.

\bibitem[Chakravarthi et~al.(2020{\natexlab{a}})Chakravarthi, Jose, Suryawanshi, Sherly, and McCrae]{chakravarthi-etal-2020-sentiment}
Bharathi~Raja Chakravarthi, Navya Jose, Shardul Suryawanshi, Elizabeth Sherly, and John~Philip McCrae.
\newblock A sentiment analysis dataset for code-mixed {M}alayalam-{E}nglish.
\newblock In Dorothee Beermann, Laurent Besacier, Sakriani Sakti, and Claudia Soria (eds.), \emph{Proceedings of the 1st Joint Workshop on Spoken Language Technologies for Under-resourced languages (SLTU) and Collaboration and Computing for Under-Resourced Languages (CCURL)}, pp.\  177--184, Marseille, France, May 2020{\natexlab{a}}. European Language Resources association.
\newblock ISBN 979-10-95546-35-1.
\newblock URL \url{https://aclanthology.org/2020.sltu-1.25/}.

\bibitem[Chakravarthi et~al.(2020{\natexlab{b}})Chakravarthi, Muralidaran, Priyadharshini, and McCrae]{chakravarthi-etal-2020-corpus}
Bharathi~Raja Chakravarthi, Vigneshwaran Muralidaran, Ruba Priyadharshini, and John~Philip McCrae.
\newblock Corpus creation for sentiment analysis in code-mixed {T}amil-{E}nglish text.
\newblock In Dorothee Beermann, Laurent Besacier, Sakriani Sakti, and Claudia Soria (eds.), \emph{Proceedings of the 1st Joint Workshop on Spoken Language Technologies for Under-resourced languages (SLTU) and Collaboration and Computing for Under-Resourced Languages (CCURL)}, pp.\  202--210, Marseille, France, May 2020{\natexlab{b}}. European Language Resources association.
\newblock ISBN 979-10-95546-35-1.
\newblock URL \url{https://aclanthology.org/2020.sltu-1.28/}.

\bibitem[Do{\u{g}}ru{\"o}z et~al.(2021)Do{\u{g}}ru{\"o}z, Sitaram, Bullock, and Toribio]{dogruoz-etal-2021-survey}
A.~Seza Do{\u{g}}ru{\"o}z, Sunayana Sitaram, Barbara~E. Bullock, and Almeida~Jacqueline Toribio.
\newblock A survey of code-switching: Linguistic and social perspectives for language technologies.
\newblock In Chengqing Zong, Fei Xia, Wenjie Li, and Roberto Navigli (eds.), \emph{Proceedings of the 59th Annual Meeting of the Association for Computational Linguistics and the 11th International Joint Conference on Natural Language Processing (Volume 1: Long Papers)}, pp.\  1654--1666, Online, August 2021. Association for Computational Linguistics.
\newblock \doi{10.18653/v1/2021.acl-long.131}.
\newblock URL \url{https://aclanthology.org/2021.acl-long.131/}.

\bibitem[Foroutan et~al.(2025{\natexlab{a}})Foroutan, Saydaliev, Kim, and Bosselut]{conlid}
Negar Foroutan, Jakhongir Saydaliev, Ye~Eun Kim, and Antoine Bosselut.
\newblock Conlid: Supervised contrastive learning for low-resource language identification, 2025{\natexlab{a}}.
\newblock URL \url{https://arxiv.org/abs/2506.15304}.

\bibitem[Foroutan et~al.(2025{\natexlab{b}})Foroutan, Saydaliev, Kim, and Bosselut]{foroutan2025conlidsupervisedcontrastivelearning}
Negar Foroutan, Jakhongir Saydaliev, Ye~Eun Kim, and Antoine Bosselut.
\newblock Conlid: Supervised contrastive learning for low-resource language identification, 2025{\natexlab{b}}.
\newblock URL \url{https://arxiv.org/abs/2506.15304}.

\bibitem[Goswami et~al.(2024)Goswami, Thilagan, North, Malmasi, and Zampieri]{goswami-etal-2024-native}
Dhiman Goswami, Sharanya Thilagan, Kai North, Shervin Malmasi, and Marcos Zampieri.
\newblock Native language identification in texts: A survey.
\newblock In Kevin Duh, Helena Gomez, and Steven Bethard (eds.), \emph{Proceedings of the 2024 Conference of the North American Chapter of the Association for Computational Linguistics: Human Language Technologies (Volume 1: Long Papers)}, pp.\  3149--3160, Mexico City, Mexico, June 2024. Association for Computational Linguistics.
\newblock \doi{10.18653/v1/2024.naacl-long.173}.
\newblock URL \url{https://aclanthology.org/2024.naacl-long.173/}.

\bibitem[Goyal et~al.(2022)Goyal, Gao, Chaudhary, Chen, Wenzek, Ju, Krishnan, Ranzato, Guzm{\'a}n, and Fan]{goyal-etal-2022-flores}
Naman Goyal, Cynthia Gao, Vishrav Chaudhary, Peng-Jen Chen, Guillaume Wenzek, Da~Ju, Sanjana Krishnan, Marc{'}Aurelio Ranzato, Francisco Guzm{\'a}n, and Angela Fan.
\newblock The {F}lores-101 evaluation benchmark for low-resource and multilingual machine translation.
\newblock \emph{Transactions of the Association for Computational Linguistics}, 10:\penalty0 522--538, 2022.
\newblock \doi{10.1162/tacl_a_00474}.
\newblock URL \url{https://aclanthology.org/2022.tacl-1.30}.

\bibitem[Hamed et~al.(2022)Hamed, Habash, Abdennadher, and Vu]{hamed-etal-2022-arzen}
Injy Hamed, Nizar Habash, Slim Abdennadher, and Ngoc~Thang Vu.
\newblock {A}rz{E}n-{ST}: A three-way speech translation corpus for code-switched {E}gyptian {A}rabic-{E}nglish.
\newblock In Houda Bouamor, Hend Al-Khalifa, Kareem Darwish, Owen Rambow, Fethi Bougares, Ahmed Abdelali, Nadi Tomeh, Salam Khalifa, and Wajdi Zaghouani (eds.), \emph{Proceedings of the Seventh Arabic Natural Language Processing Workshop (WANLP)}, pp.\  119--130, Abu Dhabi, United Arab Emirates (Hybrid), December 2022. Association for Computational Linguistics.
\newblock \doi{10.18653/v1/2022.wanlp-1.12}.
\newblock URL \url{https://aclanthology.org/2022.wanlp-1.12/}.

\bibitem[Hamed et~al.(2024)Hamed, Eryani, Palfreyman, and Habash]{hamed2024zaebucspokenmultilingualmultidialectalarabicenglish}
Injy Hamed, Fadhl Eryani, David Palfreyman, and Nizar Habash.
\newblock Zaebuc-spoken: A multilingual multidialectal arabic-english speech corpus, 2024.
\newblock URL \url{https://arxiv.org/abs/2403.18182}.

\bibitem[Jauhiainen et~al.(2019)Jauhiainen, Lui, Zampieri, Baldwin, and Lindén]{Jauhiainen2019}
Tommi Jauhiainen, Marco Lui, Marcos Zampieri, Timothy Baldwin, and Krister Lindén.
\newblock Automatic language identification in texts: A survey.
\newblock Technical report, 2019.

\bibitem[Joshi et~al.(2020)Joshi, Santy, Budhiraja, Bali, and Choudhury]{joshi-etal-2020-state}
Pratik Joshi, Sebastin Santy, Amar Budhiraja, Kalika Bali, and Monojit Choudhury.
\newblock The state and fate of linguistic diversity and inclusion in the {NLP} world.
\newblock In Dan Jurafsky, Joyce Chai, Natalie Schluter, and Joel Tetreault (eds.), \emph{Proceedings of the 58th Annual Meeting of the Association for Computational Linguistics}, pp.\  6282--6293, Online, July 2020. Association for Computational Linguistics.
\newblock \doi{10.18653/v1/2020.acl-main.560}.
\newblock URL \url{https://aclanthology.org/2020.acl-main.560/}.

\bibitem[Joulin et~al.(2016)Joulin, Grave, Bojanowski, Douze, J{\'e}gou, and Mikolov]{joulin2016fasttext}
Armand Joulin, Edouard Grave, Piotr Bojanowski, Matthijs Douze, H{\'e}rve J{\'e}gou, and Tomas Mikolov.
\newblock Fasttext.zip: Compressing text classification models.
\newblock \emph{arXiv preprint arXiv:1612.03651}, 2016.

\bibitem[Joulin et~al.(2017)Joulin, Grave, Bojanowski, and Mikolov]{joulin-etal-2017-bag}
Armand Joulin, Edouard Grave, Piotr Bojanowski, and Tomas Mikolov.
\newblock Bag of tricks for efficient text classification.
\newblock In Mirella Lapata, Phil Blunsom, and Alexander Koller (eds.), \emph{Proceedings of the 15th Conference of the {E}uropean Chapter of the Association for Computational Linguistics: Volume 2, Short Papers}, pp.\  427--431, Valencia, Spain, April 2017. Association for Computational Linguistics.
\newblock URL \url{https://aclanthology.org/E17-2068/}.

\bibitem[Kargaran et~al.(2023)Kargaran, Imani, Yvon, and Schuetze]{kargaran-etal-2023-glotlid}
Amir~Hossein Kargaran, Ayyoob Imani, Fran{\c{c}}ois Yvon, and Hinrich Schuetze.
\newblock {G}lot{LID}: Language identification for low-resource languages.
\newblock In Houda Bouamor, Juan Pino, and Kalika Bali (eds.), \emph{Findings of the Association for Computational Linguistics: EMNLP 2023}, pp.\  6155--6218, Singapore, December 2023. Association for Computational Linguistics.
\newblock \doi{10.18653/v1/2023.findings-emnlp.410}.
\newblock URL \url{https://aclanthology.org/2023.findings-emnlp.410/}.

\bibitem[Kargaran et~al.(2024)Kargaran, Yvon, and Schuetze]{kargaran-etal-2024-masklid}
Amir~Hossein Kargaran, Fran{\c{c}}ois Yvon, and Hinrich Schuetze.
\newblock {M}ask{LID}: Code-switching language identification through iterative masking.
\newblock In Lun-Wei Ku, Andre Martins, and Vivek Srikumar (eds.), \emph{Proceedings of the 62nd Annual Meeting of the Association for Computational Linguistics (Volume 2: Short Papers)}, pp.\  459--469, Bangkok, Thailand, August 2024. Association for Computational Linguistics.
\newblock \doi{10.18653/v1/2024.acl-short.43}.
\newblock URL \url{https://aclanthology.org/2024.acl-short.43/}.

\bibitem[Khanuja et~al.(2020)Khanuja, Dandapat, Srinivasan, Sitaram, and Choudhury]{khanuja-etal-2020-gluecos}
Simran Khanuja, Sandipan Dandapat, Anirudh Srinivasan, Sunayana Sitaram, and Monojit Choudhury.
\newblock {GLUEC}o{S}: An evaluation benchmark for code-switched {NLP}.
\newblock In Dan Jurafsky, Joyce Chai, Natalie Schluter, and Joel Tetreault (eds.), \emph{Proceedings of the 58th Annual Meeting of the Association for Computational Linguistics}, pp.\  3575--3585, Online, July 2020. Association for Computational Linguistics.
\newblock \doi{10.18653/v1/2020.acl-main.329}.
\newblock URL \url{https://aclanthology.org/2020.acl-main.329/}.

\bibitem[Leong et~al.(2022)Leong, Nemecek, Mansdorfer, Filighera, Owodunni, and Whitenack]{leong-etal-2022-bloom}
Colin Leong, Joshua Nemecek, Jacob Mansdorfer, Anna Filighera, Abraham Owodunni, and Daniel Whitenack.
\newblock Bloom library: Multimodal datasets in 300+ languages for a variety of downstream tasks.
\newblock In Yoav Goldberg, Zornitsa Kozareva, and Yue Zhang (eds.), \emph{Proceedings of the 2022 Conference on Empirical Methods in Natural Language Processing}, pp.\  8608--8621, Abu Dhabi, United Arab Emirates, December 2022. Association for Computational Linguistics.
\newblock \doi{10.18653/v1/2022.emnlp-main.590}.
\newblock URL \url{https://aclanthology.org/2022.emnlp-main.590/}.

\bibitem[Lovenia et~al.(2022)Lovenia, Cahyawijaya, Winata, Xu, Yan, Liu, Frieske, Yu, Dai, Barezi, et~al.]{lovenia2022ascend}
Holy Lovenia, Samuel Cahyawijaya, Genta~Indra Winata, Peng Xu, Xu~Yan, Zihan Liu, Rita Frieske, Tiezheng Yu, Wenliang Dai, Elham~J Barezi, et~al.
\newblock Ascend: A spontaneous chinese-english dataset for code-switching in multi-turn conversation.
\newblock In \emph{Proceedings of the 13th Language Resources and Evaluation Conference (LREC)}, 2022.

\bibitem[Lui \& Baldwin(2012)Lui and Baldwin]{lui-baldwin-2012-langid}
Marco Lui and Timothy Baldwin.
\newblock langid.py: An off-the-shelf language identification tool.
\newblock In Min Zhang (ed.), \emph{Proceedings of the {ACL} 2012 System Demonstrations}, pp.\  25--30, Jeju Island, Korea, July 2012. Association for Computational Linguistics.
\newblock URL \url{https://aclanthology.org/P12-3005/}.

\bibitem[Martins \& Silva(2005)Martins and Silva]{lid_web}
Bruno Martins and M\'{a}rio~J. Silva.
\newblock Language identification in web pages.
\newblock In \emph{Proceedings of the 2005 ACM Symposium on Applied Computing}, SAC '05, pp.\  764–768, New York, NY, USA, 2005. Association for Computing Machinery.
\newblock ISBN 1581139640.
\newblock \doi{10.1145/1066677.1066852}.
\newblock URL \url{https://doi.org/10.1145/1066677.1066852}.

\bibitem[Nguyen \& Bryant(2020)Nguyen and Bryant]{nguyen-bryant-2020-canvec}
Li~Nguyen and Christopher Bryant.
\newblock {C}an{VEC} - the canberra {V}ietnamese-{E}nglish code-switching natural speech corpus.
\newblock In Nicoletta Calzolari, Fr{\'e}d{\'e}ric B{\'e}chet, Philippe Blache, Khalid Choukri, Christopher Cieri, Thierry Declerck, Sara Goggi, Hitoshi Isahara, Bente Maegaard, Joseph Mariani, H{\'e}l{\`e}ne Mazo, Asuncion Moreno, Jan Odijk, and Stelios Piperidis (eds.), \emph{Proceedings of the Twelfth Language Resources and Evaluation Conference}, pp.\  4121--4129, Marseille, France, May 2020. European Language Resources Association.
\newblock ISBN 979-10-95546-34-4.
\newblock URL \url{https://aclanthology.org/2020.lrec-1.507/}.

\bibitem[Panich et~al.(2015)Panich, Conrad, and Mauve]{Panich2015ComparisonOL}
Leonid Panich, Stefan Conrad, and Martin Mauve.
\newblock Comparison of language identification techniques.
\newblock 2015.
\newblock URL \url{https://api.semanticscholar.org/CorpusID:62841821}.

\bibitem[Patwa et~al.(2020)Patwa, Aguilar, Kar, Pandey, PYKL, Gamb{\"a}ck, Chakraborty, Solorio, and Das]{patwa-etal-2020-semeval}
Parth Patwa, Gustavo Aguilar, Sudipta Kar, Suraj Pandey, Srinivas PYKL, Bj{\"o}rn Gamb{\"a}ck, Tanmoy Chakraborty, Thamar Solorio, and Amitava Das.
\newblock {S}em{E}val-2020 task 9: Overview of sentiment analysis of code-mixed tweets.
\newblock In Aurelie Herbelot, Xiaodan Zhu, Alexis Palmer, Nathan Schneider, Jonathan May, and Ekaterina Shutova (eds.), \emph{Proceedings of the Fourteenth Workshop on Semantic Evaluation}, pp.\  774--790, Barcelona (online), December 2020. International Committee for Computational Linguistics.
\newblock \doi{10.18653/v1/2020.semeval-1.100}.
\newblock URL \url{https://aclanthology.org/2020.semeval-1.100/}.

\bibitem[Prioleau \& Aryal(2023)Prioleau and Aryal]{tokenlevel_lid}
Howard Prioleau and Saurav~K. Aryal.
\newblock Benchmarking current state-of-the-art transformer models on token level language identification and language pair identification.
\newblock In \emph{2023 International Conference on Computational Science and Computational Intelligence (CSCI)}, pp.\  193--199, 2023.
\newblock \doi{10.1109/CSCI62032.2023.00036}.

\bibitem[Sabty et~al.(2021)Sabty, Mesabah, {\c{C}}etino{\u{g}}lu, and Abdennadher]{sabty2021language}
Caroline Sabty, Islam Mesabah, {\"O}zlem {\c{C}}etino{\u{g}}lu, and Slim Abdennadher.
\newblock Language identification of intra-word code-switching for arabic--english.
\newblock \emph{Array}, 12:\penalty0 100104, 2021.
\newblock \doi{10.1016/j.array.2021.100104}.

\bibitem[Srivastava \& Singh(2021)Srivastava and Singh]{srivastava2021hingedatasetgenerationevaluation}
Vivek Srivastava and Mayank Singh.
\newblock Hinge: A dataset for generation and evaluation of code-mixed hinglish text, 2021.
\newblock URL \url{https://arxiv.org/abs/2107.03760}.

\bibitem[Sterner \& Teufel(2023)Sterner and Teufel]{sterner2023tongueswitcher}
Igor Sterner and Simone Teufel.
\newblock Tongueswitcher: Fine-grained identification of german-english code-switching.
\newblock In \emph{Sixth Workshop on Computational Approaches to Linguistic Code-Switching}. Empirical Methods in Natural Language Processing, 2023.

\bibitem[Thoma(2018)]{thoma2018wilibenchmarkdatasetwritten}
Martin Thoma.
\newblock The wili benchmark dataset for written language identification, 2018.
\newblock URL \url{https://arxiv.org/abs/1801.07779}.

\bibitem[Tiedemann(2020)]{tiedemann-2020-tatoeba}
J{\"o}rg Tiedemann.
\newblock The tatoeba translation challenge {--} realistic data sets for low resource and multilingual {MT}.
\newblock In Lo{\"i}c Barrault, Ond{\v{r}}ej Bojar, Fethi Bougares, Rajen Chatterjee, Marta~R. Costa-juss{\`a}, Christian Federmann, Mark Fishel, Alexander Fraser, Yvette Graham, Paco Guzman, Barry Haddow, Matthias Huck, Antonio~Jimeno Yepes, Philipp Koehn, Andr{\'e} Martins, Makoto Morishita, Christof Monz, Masaaki Nagata, Toshiaki Nakazawa, and Matteo Negri (eds.), \emph{Proceedings of the Fifth Conference on Machine Translation}, pp.\  1174--1182, Online, November 2020. Association for Computational Linguistics.
\newblock URL \url{https://aclanthology.org/2020.wmt-1.139/}.

\bibitem[Winata et~al.(2023)Winata, Aji, Yong, and Solorio]{winata-etal-2023-decades}
Genta Winata, Alham~Fikri Aji, Zheng~Xin Yong, and Thamar Solorio.
\newblock The decades progress on code-switching research in {NLP}: A systematic survey on trends and challenges.
\newblock In Anna Rogers, Jordan Boyd-Graber, and Naoaki Okazaki (eds.), \emph{Findings of the Association for Computational Linguistics: ACL 2023}, pp.\  2936--2978, Toronto, Canada, July 2023. Association for Computational Linguistics.
\newblock \doi{10.18653/v1/2023.findings-acl.185}.
\newblock URL \url{https://aclanthology.org/2023.findings-acl.185/}.

\bibitem[Yirmibe{\c{s}}o{\u{g}}lu \& Eryi{\u{g}}it(2018)Yirmibe{\c{s}}o{\u{g}}lu and Eryi{\u{g}}it]{yirmibesoglu-eryigit-2018-detecting}
Zeynep Yirmibe{\c{s}}o{\u{g}}lu and G{\"u}l{\c{s}}en Eryi{\u{g}}it.
\newblock Detecting code-switching between {T}urkish-{E}nglish language pair.
\newblock In Wei Xu, Alan Ritter, Tim Baldwin, and Afshin Rahimi (eds.), \emph{Proceedings of the 2018 {EMNLP} Workshop W-{NUT}: The 4th Workshop on Noisy User-generated Text}, pp.\  110--115, Brussels, Belgium, November 2018. Association for Computational Linguistics.
\newblock \doi{10.18653/v1/W18-6115}.
\newblock URL \url{https://aclanthology.org/W18-6115/}.

\bibitem[Zhang \& Eickhoff(2024)Zhang and Eickhoff]{zhang-eickhoff-2024-crocosum}
Ruochen Zhang and Carsten Eickhoff.
\newblock {C}ro{C}o{S}um: A benchmark dataset for cross-lingual code-switched summarization.
\newblock In \emph{Proceedings of the 2024 Joint International Conference on Computational Linguistics, Language Resources and Evaluation (LREC-COLING 2024)}, pp.\  4113--4126, Torino, Italia, May 2024. ELRA and ICCL.
\newblock URL \url{https://aclanthology.org/2024.lrec-main.367/}.

\bibitem[Zubiaga et~al.(2016)Zubiaga, Vicente, Gamallo, Pichel, Alegria, Aranberri, Ezeiza, and Fresno]{tweetlid}
Arkaitz Zubiaga, Iñaki~San Vicente, Pablo Gamallo, José~Ramom Pichel, Iñaki Alegria, Nora Aranberri, Aitzol Ezeiza, and Víctor Fresno.
\newblock Tweetlid: a benchmark for tweet language identification.
\newblock \emph{Language Resources and Evaluation}, 50\penalty0 (4):\penalty0 729--766, 2016.
\newblock ISSN 1574-0218.
\newblock \doi{10.1007/s10579-015-9317-4}.
\newblock URL \url{https://doi.org/10.1007/s10579-015-9317-4}.

\end{thebibliography}
